\documentclass[cameraready]{Interspeech}

\title{Accent Vector: Controllable Accent Manipulation for Multilingual TTS Without Accented Data}

\author[affiliation={1}, equalcontribution, orcid=0009-0009-9214-1743]{Thanathai}{Lertpetchpun}
\author[affiliation={2}, equalcontribution, orcid=0009-0006-4866-5011]{Thanapat}{Trachu}
\author[affiliation={1}, orcid=0009-0009-3033-2042]{Jihwan}{Lee}
\author[affiliation={1}, orcid=0000-0002-2053-9068]{Tiantian}{Feng}
\author[affiliation={3}, orcid=0000-0003-3319-5871]{\\ Dani}{Byrd}
\author[affiliation={1,2,3}, orcid=0000-0002-1052-6204]{Shrikanth}{Narayanan}

\address{
    $^1$Signal Analysis and Interpretation Lab, University of Southern California, USA \\
    $^2$Thomas Lord Department of Computer Science, University of Southern California, USA \\
    $^3$Department of Linguistics, University of Southern California 
}

\email{lertpetc@usc.edu}

\keywords{Text-To-Speech model, Controllable Speech Synthesis, Accented Speech Generation}

\usepackage{comment}
\usepackage{amssymb}
\usepackage{pifont}
\newcommand{\cmark}{\ding{51}}
\newcommand{\xmark}{\ding{55}}
\usepackage{stackengine}
\usepackage{cite}
\newcommand{\improvement}[1]{{\scriptsize \color{gray} (#1)}}

\begin{document}

\maketitle

\begin{abstract}
 
Accent is an integral part of society, reflecting multiculturalism and shaping how individuals express identity. The majority of English speakers are non-native (L2) speakers, yet current Text-To-Speech (TTS) systems primarily model American-accented English due limited accented data. We propose \textit{Accent Vector}, a controllable representation that enables accent manipulation in multilingual TTS without requiring accented training data. \textit{Accent Vector} is derived by fine-tuning a TTS system on native speech of a different language (i.e. non-English) and computing task vectors capturing accent characteristics (i.e. in English).  By scaling and interpolating the vector, we achieve fine-grained control over accent strength and generate mixed-accent speech. In addition, it generalizes beyond English, enabling accent control across multiple languages. Objective and human evaluations confirm the effectiveness of Accent Vector for fine-grained and compositional accent control.

\end{abstract}

\section{Introduction}

English is spoken by approximately one-fifth of the world’s population, yet only about one-quarter of these speakers have acquired it as a first language (L1)~\cite{crystal2003english,ethno2024english}. The language exhibits a wide range of regional and global accents across both first-language (L1) and second-language (L2) speakers~\cite{wells1982accents}. Accent, in this sense, is not a deviation from a norm but an index of variation in linguistic backgrounds, which may reflect early phonological acquisition as well as subsequent language contact. Accent also reflects differences in segmental inventory, intonation pattern, stress placement, and vowel quality. For example, L2 English speakers may retain timing patterns or vowel distinctions influenced by their first language, producing speech that is fully intelligible but characterized by distinct rhythmic and pronunciation patterns relative to many L1 English-speaking varieties.

Despite these global variations in English varieties, most current text-to-speech (TTS) models are predominantly trained on mainstream English~\cite{chen2025f5,wang2023neural,chen2024vall,le2023voicebox}. This imbalance in training knowledge is largely driven by the availability of relevant datasets, where large-scale, high-quality, and well-annotated speech datasets are far more accessible for American English~\cite{panayotov2015librispeech, kang2024libriheavy, libritts_r, kawamura2024libritts} than for other regional L1 or L2 English varieties~\cite{zhao2018l2, yamagishi2019cstr, ardila2020common}. While a small number of accented L2 English speech datasets do exist, they are limited in both scale and coverage, and are generally insufficient for training high-quality TTS systems. This data imbalance in training will eventually impact speech synthesis performance, with synthesis quality disproportionately higher for the dominant accent but substantially lower for underrepresented English varieties.



Recent work has explored accent generation in TTS without relying on large-scale accented speech datasets. Existing approaches~\cite{onda2025prosodically,inoue2025macst} include conditioning multilingual TTS systems through text transliteration or steering model components toward a target language. These methods demonstrate that accent characteristics can be induced from native-language resources alone. However, they typically provide limited control over accent strength and often focus on specific linguistic aspects, such as pronunciation mapping or duration modeling. As a result, achieving scalable and fine-grained control over accent variation remains an open challenge.


To enhance model controllability, Ilharco et al.~\cite{taskvector} introduced task vectors, a representation of task-specific parameter shifts in pretrained models. Recent work in speech processing has leveraged task vectors to model controllable speech attributes, including emotion and other paralinguistic factors~\cite{murata2025speaker, feng2025task}. Inspired by this idea, we propose the Accent Vector, a controllable parameter shift that enables explicit regulation of accent strength in multilingual speech synthesis. Our approach obtains the Accent Vector by fine-tuning a multilingual TTS model with the objective of reconstructing and condition on reference speech from a target language (e.g., Spanish) to get the target accent (e.g., Spanish Accent). The target accent may correspond either to a regional variety of the same language (e.g., British English) or to the influence of a different native language (e.g., Spanish-accented English). Intuitively, shifting the model parameters toward the target accent direction alters not only segmental characteristics, such as phoneme realization, but also suprasegmental properties—including temporal structure (duration), rhythm, and prosodic patterns—since these attributes are encoded implicitly in the model weights.

\begin{table*}[h!]
\vspace{-10pt}
\centering
\caption{Comparison between accented TTS models.}
\label{tab:related_work}
\footnotesize
\begin{tabular}{lccccc}
\toprule
\multirow{2}{*}{\textbf{Approach}} & Requiring  & Explicit & Zero-shot Speaker & Cross-Accent & Multilingual\\
 & Accented Dataset & Accent Control & Adaptation & Mixing & Support\\
\midrule
Xinyuan et al. \cite{xinyuan2025scalable} & Required & \xmark & \cmark & \xmark & \xmark\\
Onda et al. \cite{onda2025prosodically} & Not Required & \xmark & \cmark & \xmark & \xmark \\
MacST \cite{inoue2025macst} & Not Required & \xmark & \cmark & \xmark & \xmark\\
Lertpetchpun et at. \cite{lertpetchpun2026quantifying} & Not Required & Phoneme-level & \xmark &\xmark & \xmark\\
Accent Vector (Ours) & Not Required & Weight Scaling & \cmark & \cmark & \cmark \\
\bottomrule
\end{tabular}
\end{table*}

In the case of English, for first-language (L1) or same-language accents, we fine-tune the model using speech from specific regional varieties (e.g., British English). For second-language (L2) accents, we instead fine-tune the model on speech from the speaker in the native language (e.g., Spanish, Hindi, or French). This approach removes the need for large-scale accented English speech datasets and allows us to exploit existing high-resource speech corpora in the target language. We would highlight that our method can generalize naturally beyond generating accented English speech, and our experiments on Spanish, Mandarin, French, and German show that the same learning paradigm can effectively produce accent variations in a wide range of languages. Moreover, Accent Vectors can be combined to model mixed accents, capturing cases where non-native English speakers exhibit influences from both their native language and prolonged exposure to another accent, such as British English. This is particularly relevant for capturing the speech of individuals who spend considerable time in two different linguistic environments/countries.

In summary, we propose \textit{Accent Vector}, a simple yet effective framework for controllable accented speech synthesis. \textit{Accent Vector} 1) enables accented TTS without the need for accent-specific speech datasets, 2) enables explicit control over accent strength, 3) generalizes to other languages beyond English, and 4) supports the composition of multiple accents to model mixed-accent speech. The synthesized samples are available here~\footnote{https://anonymous.4open.science/w/Accent-Vector-Audio-Samples-0EAA/}.


\section{Background and Related work}

\subsection{Task Vector}
\begin{figure*}[h!]
    \centering
    \includegraphics[width=.8\textwidth]{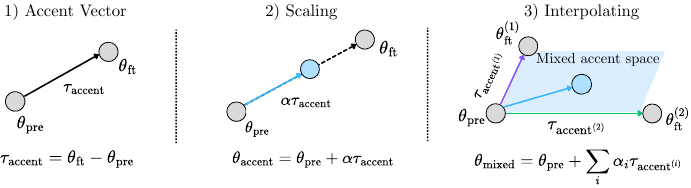}
    \vspace{-3mm}
    \caption{Accent Vector Arithmetic. The figure illustrates the computation of the Accent Vector, its scaling for accent strength control, and the interpolation of multiple Accent Vectors for mixed-accent synthesis.}
    \vspace{-5mm}
    \label{fig:diagram}
\end{figure*}

Task vectors~\cite{taskvector} are motivated by the observation that the parameter space of large pretrained models exhibits approximate linearity~\cite{ilharco2022patching}. Concretely, a task vector is defined as the difference between the parameters of a model fine-tuned on a specific task and those of its corresponding pretrained model. Under this assumption, the task vector can be interpreted as the direction in the parameter space such that moving the model parameters along this direction improves performance on the desired downstream applications. It is critical to highlight that this formulation comprises compositionality, and since different directions are approximately linear, multiple task vectors can be combined by vector addition. This allows the model to integrate knowledge from different fine-tuning datasets and downstream applications, enabling improved performance across multiple tasks. Based on this formulation, we treat accent adaptation as the parameter shift, and we define the Accent Vector that encodes accent-specific speech characteristics. For example, as shown in Figure~\ref{fig:diagram}, scaling the magnitude of the Accent Vector provides explicit control over the accent strength, while the addition of multiple Accent Vectors introduces the composition of multiple accents in speech synthesis.

\subsection{XTTS}
\label{sec:xtts_background}
XTTS~\cite{casanova2024xtts} is a multilingual zero-shot text-to-speech (ZS-TTS) model that supports 17 languages. The model consists of three main components: 1) a Vector Quantized Variational Autoencoder (VQ-VAE)~\cite{betker2023better}, 2) an encoder, and 3) a decoder. The VQ-VAE takes the ground-truth spectrogram as input and converts it into discrete acoustic codes, which serve as training targets for the encoder. The encoder receives the text transcript, reference speech, and a language identification token. The text is tokenized using Byte-Pair Encoding (BPE)~\cite{gage1994new}, while the reference speech is processed through attention-based layers to capture speaker and acoustic characteristics. The encoder is trained to predict the discrete codes produced by the VQ-VAE, using a cross-entropy loss. The latent representation generated by the encoder is then passed to the decoder, which synthesizes speech corresponding to the input transcript and reference speech. Then, it was trained with a mel-spectrogram reconstruction loss alongside adversarial GAN loss and a speaker consistency loss. In this work, we adopt XTTS as the backbone model and apply Low-Rank Adaptation (LoRA)~\cite{hu2022lora} for fine-tuning, which reduces the number of trainable parameters and mitigates overfitting and catastrophic forgetting~\cite{biderman2024lora}.

\subsection{Accented TTS}

Recent advancements in Text-to-Speech (TTS) and voice conversion have made significant strides in generating and modifying accents to address the challenge of lacking sufficient accented speech datasets.  The overview comparison between accented TTS models is shown in Table~\ref{tab:related_work}. 

One line of work conditions TTS systems on accent labels. For example, \cite{xinyuan2025scalable} propose training a TTS model using geographically grounded labels, allowing the system to generate speech conditioned on a target accent. While effective, this approach depends on the availability of fine-grained geographical metadata, which is not consistently provided across large-scale speech corpora, limiting scalability.

To reduce reliance on accented speech datasets, MacST~\cite{inoue2025macst} introduces accents through text transliteration. Large Language Models (LLMs) generate transliterated text reflecting target-language pronunciation patterns, which is then synthesized using a multilingual TTS model. This approach avoids the need for accented recordings but produces a fixed accent realization determined by the transliteration process.

Taking a linguistically grounded perspective, Lertpetchpun et al.~\cite{lertpetchpun2026quantifying} apply explicit phonetic transformation rules to modify pronunciation before synthesis. Their results show that systematic phoneme substitutions and deletions can effectively transform one accent into another. This rule-based method provides coarse control over accent strength by varying the number of applied rules. However, its scalability is limited, as new accent pairings require manual rule design and linguistic expertise.

More recently, Onda et al.~\cite{onda2025prosodically} propose inducing accents using only native speech corpora. Their method leverages self-supervised representations, K-means clustering, and a duration predictor trained on the target language to simulate accent-specific prosody. While this approach reduces dependence on manual annotation, it has been validated only on Japanese-accented English. Additionally, it primarily models durational differences, leaving segmental and subsegmental variations largely unaddressed.

Our work shares the goal of eliminating the need for accent-specific English datasets but differs in several key aspects. We introduce the \textit{Accent Vector}, a controllable parameter shift that enables explicit regulation of accent strength at the model level. Unlike transliteration- or duration-focused approaches, our method provides unified control over both segmental and suprasegmental features—including phoneme realization, temporal structure (duration), and prosodic patterns—through direct parameter manipulation. In addition to fine-grained intensity control, the framework supports linear composition of multiple accents and demonstrates effectiveness across a diverse set of native languages and accents, including German, Hindi, English, French, Spanish, and Mandarin.


\begin{figure}[h!]
    \centering
    \includegraphics[width=.40\textwidth]{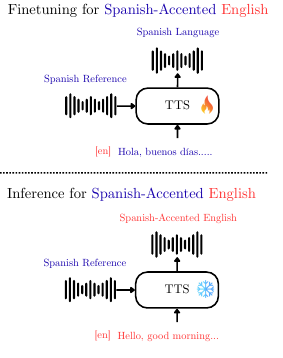}
    \vspace{-3mm}
    \caption{Accent Vector Framework. The top panel illustrates the fine-tuning procedure, and the bottom panel shows the inference process for generating accented speech. During inference, a language ID token (e.g., [en]) is concatenated with the transcript and provided as input to the model.}
    \vspace{-5mm}
    \label{fig:training_diagram}
\end{figure}

\section{Method}

In this section, we present the proposed \textit{Accent Vector} framework. First, we describe the fine-tuning procedure of the multilingual TTS model. Next, we detail the formulation and extraction of the Accent Vector, which enables explicit control over accent strength as well as the interpolation of multiple accents. Finally, we outline the inference procedure for generating accented speech using the obtained Accent Vector.


\subsection{Fine-tuning Procedure}
Our approach builds on the pretrained multilingual TTS model XTTS2~\cite{xtts}, which generates speech conditioned on a language ID, a text transcript, and a reference speech signal. In the accented speech synthesis, two linguistic dimensions are involved: a base language (the language to be synthesized) and a target accent. The target accent may correspond either to a regional variant of the same language (e.g., British English) or to a different language (e.g., Spanish-accented English). In the latter case, English is treated as the base language and Spanish as the target accent language. We fine-tune XTTS2 following the procedure described in~\cite{xtts}. During training, the model is conditioned on the base language ID while using transcripts and reference speech from native speakers of the target accent. Figure~\ref{fig:training_diagram} illustrates the exemplar process of fine-tuning Spanish-accented English. Specifically, we provide Spanish reference speech and Spanish transcripts while setting the language ID to English. The model is optimized using the same training objectives as in~\cite{xtts}, with Spanish speech as the target output.


\subsection{Obtaining Accent Vector}
After fine-tuning, we obtain the adapted model parameters $\theta_\text{ft}$. The Accent Vector is then computed as the difference between the fine-tuned model $\theta_\text{ft}$ and the pretrained model $\theta_\text{pre}$:

\begin{equation}
\tau_{\text{accent}} = \theta_\text{ft} - \theta_\text{pre}.
\end{equation}

Note that the finetuned model is obtained through the Low-Rank Adaptation. Then, the Accent Vector is equivalent to
\begin{align}
    \tau_{\text{accent}} &= \theta_\text{ft} - \theta_\text{pre} \\
    &= (\theta_\text{pre} + \theta_{LoRa}) - \theta_\text{pre} = \theta_{LoRa},
\end{align}
where $\theta_{LoRa}$ is the weight of the LoRa. This Accent Vector represents a direction in the model parameter space that encodes accent-specific characteristics. Importantly, this process is not limited to English as the base language. The same procedure can be applied to any base language supported by the multilingual TTS model. For example, by selecting Mandarin or Spanish as the base language and British English as the target accent, the model can be fine-tuned to produce British-accented Mandarin or British-accented Spanish speech. The overall procedure remains unchanged, requiring only a swap of the base language while preserving the same fine-tuning and Accent Vector extraction process.


\subsection{Accent Vector Arithmetic}
At inference time, we apply Accent Vector arithmetic to control the presence and strength of the accent. Starting from the pretrained model parameters $\theta_\text{ft}$, we construct an accent-modified model by adding the Accent Vector scaled by a control coefficient $\alpha$:

\begin{equation}
\label{eq:coefficient}
\theta_\text{accent} =\theta_\text{pre}+ \alpha \cdot \tau_{\text{accent}}.
\end{equation}

The scalar $\alpha$ provides continuous control over accent strength, where larger values correspond to stronger accent characteristics. During inference, the model is conditioned on the base language ID and transcripts in the base language. For example, in the case of Spanish-accented English, the transcript is provided in English, while the accent is introduced solely through the modified model parameters. All other inference settings, including reference speech conditioning, remain consistent with those used during fine-tuning.

One advantage of Accent Vectors is that they can be linearly composed to model mixed accents. Given multiple Accent Vectors \( AV_{\text{accent}}^{(i)} \), each corresponding to a distinct accent or language, we construct a mixed accent representation by weighted addition. Specifically, the mixed Accent Vector is defined as

\vspace{-2mm}
\begin{equation}
\label{eq:mixed_accent_1}
\tau_{\text{interpolated}} = \sum_{i=1}^{N} \alpha_i \cdot \tau_{{\text{accent}}^{(i)}} ,
\end{equation}
\vspace{-2mm}

where \( \alpha_i \) controls the relative contribution of the \( i \)-th accent. The parameters of the mixed-accent model are then obtained by applying the composed Accent Vector to the pretrained model,
\vspace{-2mm}
\begin{equation}
\label{eq:mixed_accent_2}
\theta_\text{interpolated}= \theta_\text{pre} + \tau_{\text{interpolated}} .
\end{equation}
This enables flexible modeling of speakers exhibiting multiple accent influences. For example, a non-native English speaker who has lived in the UK may exhibit speech characteristics shaped by both their native language and British English. By adjusting the coefficients \( \alpha_i \), our approach allows nuance, continuous control over the strength and balance of each accent component in the synthesized speech.


\subsection{Inference Procedure}
Figure~\ref{fig:training_diagram} illustrates the inference procedure after obtaining the Accent Vector. For example, to generate Spanish-accented English, we condition the TTS system on the English language ID and English transcript, together with a Spanish reference signal (or its corresponding accent). This behavior can be attributed to the fine-tuning process on the target accent language, during which the model learns to align the phonetic and phonotactic representations of the base language with those of the target accent. Consequently, the synthesized speech preserves the linguistic content of the base language while exhibiting acoustic characteristics of the target accent. 


\section{Experimental Setup}

%
\subsection{Text-to-speech Modeling}
We use the pretrained XTTS-v2 model released on Hugging Face\footnote{\url{https://huggingface.co/coqui/XTTS-v2}} as the base model in all experiments. Fine-tuning is performed using Low-Rank Adaptation (LoRA) \cite{hu2022lora} with 16 ranks for 60,000 training steps, where LoRA is applied to all linear layers in the encoder modules as described in Section~\ref{sec:xtts_background}. By employing LoRA, the number of trainable parameters is drastically reduced from 378M, required for normal fine-tuning, to approximately 8M. We employ the Adam optimizer~\cite{kingma2014adam} with a learning rate of $3\mathrm{e}{-5}$ and follow the fine-tuning procedure described in \cite{xtts}. The experiments were conducted using one A40 for 8 GPU hours.

\subsection{Datasets}

\begin{table*}[h]
\centering
\caption{Languages and corresponding datasets. UTMOS is reported after filtering low-quality samples.}
\footnotesize
\begin{tabular}{lcccccc}
\toprule
\textbf{Language} & \textbf{Training Dataset} & \textbf{Subset} & \textbf{Duration (Hours)} & \textbf{UTMOS (Train)} & \textbf{UTMOS (Eval)} \\ 
\midrule
English & VCTK & English & 8.9 & 3.92 & 3.86 \\ 
Spanish & CommonVoice-sp & Penisular & 52.21 & 2.79 & 2.61 \\ 
Hindi & IndicVoices-R & Uttar Pradesh and Madhya Pradesh & 27.2 & 2.57 & 2.57 \\ 
German & CommonVoice-de & German-Non-NRW and -NRW Area & 375.12 & 2.93 & 2.86 \\ 
French & CommonVoice-fr & France & 337.56 & 2.63 & 2.50 \\ 
Mandarin & KeSpeech & Beijing, Northeastern & 125.53 & 2.57 & 2.68 \\ 
\bottomrule
\end{tabular}
\label{tab:languages_datasets}
\end{table*}

We use VCTK \cite{vctk}, Common Voice \cite{commonvoice}, IndicVoices-R \cite{sankar2024indicvoicesr}, and KeSpeech \cite{kespeech} as training data sources for different accents (e.g., British English) and languages (e.g., Spanish, Hindi, German, French, and Mandarin). Table~\ref{tab:languages_datasets} provides detailed statistics, data usage, and UTMOS of the training and evaluation sets for each language. As multiple dialects exist within each language, we restrict the training data to a single representative dialect per language in order to model a more consistent accent. For example, Peninsular Spanish differs substantially from Latin American Spanish~\cite{hualde2014spanish}; for the time being, we limit the Spanish data to Peninsular Spanish to represent a single coherent Spanish accent. Specifically, for English, we select the England accent from the VCTK dataset as the representative British accent. For Spanish, German, and French, we use language-specific subsets from Common Voice, selecting Peninsular Spanish, France French, and German speakers from both NRW and non-NRW regions as representative varieties. For Hindi, we use IndicVoices-R and filter the data by state, selecting speakers from regions associated with Standard Hindi. For Mandarin, we select speech from Beijing and Northeastern regions, which are commonly considered representative of Standard Mandarin. Dialect labels and regional classifications follow the annotations provided in Voxlect \cite{feng2025voxlect}.

As training high-quality TTS models requires high-quality speech recordings, we apply data cleaning to all datasets by selecting only utterances with a DNSMOS \cite{dnsmos} score higher than 3.4 following \cite{ma2025meanvc}. Meanwhile, we discard utterances with a duration less than 3 seconds, given that shorter utterances may lack sufficient information to reliably characterize accents. However, we note that even after filtering low-quality recordings, the UTMOS scores of both the training and evaluation sets remain moderate rather than high, which may limit the overall naturalness and quality of the synthesized speech. During inference, we use a total of around 4.8k transcripts from the test-clean set of LibriTTS-R~\cite{libritts_r} and randomly pair them with speech samples from the target-language dataset to generate accented speech in different target-languages. 


\section{Evaluation Metrics}
Since our goal is to generate accented speech while preserving overall speech quality, we evaluate the proposed framework from two complementary perspectives: accentedness and utility. Accentedness measures how well the synthesized speech reflects the intended accent, while utility assesses speech quality and intelligibility.

\subsection{Objective Accent Evaluation}
\label{secion:objective_evaluation}
Accent quality is assessed using two automatic approaches: an accent classification model (VoxProfile)~\cite{feng2025vox} and a Spoken Language Identification (LID) model.

We employ the narrow English-accent classification model~\footnote{\url{https://huggingface.co/tiantiaf/whisper-large-v3-narrow-accent}} from VoxProfile to measure accent characteristics. We report (1) the softmax posterior probability of the target accent label and (2) an embedding-based similarity score computed as the cosine similarity between synthesized speech embeddings and a representative accent embedding derived from real speech samples following~\cite{lertpetchpun2026quantifying}.

Evaluating cross-lingual accent transfer presents a practical challenge. Although accent classifiers exist for several languages, they are typically designed to distinguish major native dialects within a single language~\cite{feng2025voxlect} and are not trained to recognize rarer cross-lingual accent configurations (e.g., English-accented Spanish or English-accented Mandarin). Prior analysis suggests that accent classifiers primarily rely on phoneme-level cues~\cite{yang2023can}. Based on this observation, we assume that if synthesized speech exhibits phonetic characteristics associated with English (e.g., England accent), VoxProfile will assign a higher probability to the English accent label, even when the base language differs. We therefore use VoxProfile as an indirect measure of English-accent strength in non-English speech.

As this evaluation remains approximate in cross-lingual settings, we additionally incorporate an ECAPA-TDNN LID model\footnote{\url{https://huggingface.co/speechbrain/lang-id-voxlingua107-ecapa}} trained on VoxLingua107~\cite{valk2021slt} as a complementary proxy. Prior work~\cite{bafna25_interspeech} shows that LID systems frequently misclassify L2-accented speech as the speaker’s native language. Accordingly, we interpret the increased probability assigned to the accent source language by the LID model as evidence stronger accent presence.

\subsection{Speech Quality}
For utility evaluation, we report word error rate (WER), character error rate (CER), UTMOS~\cite{saeki2022utmos}, and speaker similarity. Synthesized Transcripts are obtained from using the Whisper-medium automatic speech recognition (ASR) model\footnote{\url{https://huggingface.co/openai/whisper-medium}}~\cite{radford2023robust}. We compute WER and CER only for utterances shorter than 30 seconds. Prior to evaluation, both predicted transcripts and ground-truth text are normalized using Whisper’s EnglishTextNormalizer\footnote{\url{https://github.com/openai/whisper/blob/main/whisper/normalizers/english.py}} for English, and WeTextProcessing\footnote{\url{https://github.com/wenet-e2e/WeTextProcessing}} for Mandarin. To reduce the impact of ASR artifacts, we discard samples whose inferred transcripts contain more than six words per second and whose ground-truth transcripts contain fewer than two words, following the typical utterance length distribution used in TTS pretraining. We use UTMOS~\cite{saeki2022utmos} as an automatic measure of naturalness, with scores ranging from 1 (least natural) to 5 (human-level naturalness). To evaluate speaker consistency, we compute speaker similarity using embeddings extracted from wavlm-base-plus-sv~\cite{chen2022wavlm}\footnote{\url{https://huggingface.co/microsoft/wavlm-base-plus-sv}}. Specifically, we calculate the cosine similarity between the reference and synthesized speech embeddings, where values closer to 1 indicate higher speaker similarity.

\subsection{Subjective Evaluation}
To complement the objective evaluation, we also perform a subjective evaluation by evaluating the speech samples by human panel. We focus on these three aspects: perceived accent, accent strength, and naturalness.
We recruited 16 human listeners and total 70 speech samples were evaluated. They consist of both native and fluent speakers of English, and for the non-native fluent speakers, their native languages vary from Asian and European languages. They all currently live in the US. They first were asked to choose the accent they hear, then rate how prominent the accent is from a scale of 1 to 5: [1:not at all prominent, 2:slightly prominent, 3:moderately prominent, 4:quite prominent, 5:extremely prominent]. They were also asked to rate naturalness of the speech samples in the same scale.

\section{Results and Discussion}
We evaluate the proposed Accent Vector framework across four experimental settings to demonstrate its effectiveness and controllability: 1) accented speech synthesis in English, 2) generalization to non-English base languages, including Spanish, German, and Mandarin, 3) fine-grained control over accent strength, 4) blending and controlling of two distinct accents within a single utterance.

\subsection{Multiple Accented English Speech Evaluation}
\begin{table*}[h!]
\centering
\caption{Objective evaluation on accent shifting from American English to each target accent. We report the accent prediction probabilities (Prob) and accent cosine similarities measured by VoxProfile, intelligibility (WER/CER), speaker similarity (SSIM), and naturalness (UTMOS).}
\label{tab:accent_shift}
\footnotesize
\begin{tabular}{lccccccccc}
\toprule
\multirow{2}{*}{\textbf{Target Accent}} 
& \multicolumn{2}{c}{\textbf{Target Accent} Prob. (\%)} 
& \multicolumn{2}{c}{\textbf{Target Accent} SIM. (\%)} 
& \multirow{2}{*}{WER / CER}
& \multirow{2}{*}{SSIM} 
& \multirow{2}{*}{UTMOS} 
 \\
\cmidrule(lr){2-3} \cmidrule(lr){4-5} 
& Pretrained 
& Finetuned 
& Pretrained 
& Finetuned 
& 
& 
& \\
\midrule
British (England Accent) & 23.3 & 56.7\improvement{+143.83\%} & 0.47 & 0.73 & 5.46 / 4.31 & 0.90 & 3.61 & \\
Spanish  & 15.5 & 39.7\improvement{+156.37\%} & 0.41 & 0.61 & 17.3 / 10.3 & 0.88 & 2.86 & \\
Hindi    & 2.2 & 24.2\improvement{+1021.29\%} & 0.40 & 0.64 & 10.4 / 6.5 & 0.88 & 3.10 & \\
French   & 12.2 & 23.2\improvement{+90.10\%} & 0.38 & 0.56 & 18.9 / 11.4 & 0.87 & 2.76 & \\
German   & 14.3 & 27.4\improvement{+91.14\%} & 0.50 & 0.52 & 11.0 / 6.8 & 0.86 & 3.04 & \\
Mandarin  & 27.4 & 33.8\improvement{+23.56\%} & 0.56 & 0.65 & 34.4 / 23.6 & 0.86 & 2.59 &  \\
\bottomrule
\end{tabular}
\end{table*}
\subsubsection{Accent Shift Effectiveness}
We evaluate our method on six English accents: British, Spanish, Hindi, German, French, and Mandarin. For each accent, the model is fine-tuned using speech data from the corresponding language while keeping the language identity token fixed to English. Table~\ref{tab:accent_shift} summarizes the results; we report both the accent classification probability and the accent similarity for the corresponding accent group (e.g., South Asian for Hindi-accented English). The “Pretrained” column denotes the original XTTS2 model without fine-tuning, conditioned only on the target-language reference speech during inference.

Across all six accents, our method consistently shifts the synthesized speech toward the intended target accent. Both accent probability and accent similarity increase relative to the pretrained baseline, demonstrating the effectiveness of the proposed Accent Vector. Importantly, speaker similarity remains high (around 0.9), indicating that accent manipulation preserves speaker identity while modifying accent characteristics.

\subsubsection{ASR Degradation and Accentedness}

We observe higher WER and CER for accented speech compared to standard English synthesis. This behavior is expected, as Whisper is predominantly trained on standard American English and has been shown to degrade on accented or non-native speech. Prior work~\cite{peng2024evaluating} similarly reports reduced ASR accuracy across diverse accents, reflecting inherent acoustic and lexical biases in ASR systems. In this context, increased WER and CER can partially reflect stronger accentedness. For example, pronunciation variations in Hindi L1 speakers, such as ``very'' realized as ``wery,'' may lead to systematic ASR mismatches.

\subsubsection{Mandarin Accent: Prosodic and Phonological Constraints}

Among the evaluated accents, Mandarin exhibits the smallest relative improvement in accent probability. We attribute this primarily to substantial prosodic differences between Mandarin and English. Mandarin is a tonal, syllable-timed language in which fundamental frequency variation (F0) encodes lexical meaning, whereas English is stress-timed and relies more heavily on rhythmic stress and intonation patterns~\cite{ding2021_interspeech}. These suprasegmental discrepancies likely make it more difficult to shift the pretrained model’s English-centric parameter distribution toward capturing Mandarin-specific prosodic characteristics. Adapting a model largely shaped by English rhythm and intonation to reflect Mandarin tonal structure therefore remains inherently challenging.

Mandarin-accented English also yields consistently higher WER and CER. This can be explained by phonological interference. Mandarin has a different phonemic inventory and much more restrictive syllable structure compared to English~\cite{WANG2025103168}. Substitution or omission of English phonemes results in pronunciations that deviate further from standard English, increasing ASR errors—particularly for systems trained predominantly on native English speech.

\begin{table*}[t!]
\centering
\caption{Objective evaluation on English-accented speech generation for non-English languages. We report English accent probability measured by VoxProfile, English language prediction probability by the LID model, intelligibility (WER/CER), speaker similarity (SSIM), and naturalness (UTMOS).}
\label{tab:accent_shift_non_english}
\footnotesize
\begin{tabular}{l cc cc cc c c c}
\toprule
\multirow{2}{*}{\textbf{Target Language}} 
& \multicolumn{2}{c}{\textbf{English Accent} Prob. (\%)} 
& \multicolumn{2}{c}{\textbf{English Language} Prob. (\%)}
& \multirow{2}{*}{WER / CER}
& \multirow{2}{*}{SSIM} 
& \multirow{2}{*}{UTMOS} 
 \\
\cmidrule(lr){2-3} \cmidrule(lr){4-5} 
& Pretrained 
& Finetuned 
& Pretrained
& Finetuned
& 
& 
&  \\
\midrule
Spanish  & 1.20 & 44.69 & 0.04 & 38.1 & 30.1 / 14.6 & 0.89 & 3.49 \\
German  & 8.57 & 41.57 & 1.13 & 35.1 & 34.2 / 12.6 & 0.89 & 3.55 \\
Mandarin  & 0.00 & 3.03 & 0.03 & 10.8 &  -- / 31.6 & 0.82 & 3.51  \\
\bottomrule
\end{tabular}
\end{table*}

\subsection{English-Accented Non-English Speech}

Moving on from previous section, we next investigate whether the proposed framework can induce a British (England) accent in languages other than English. To demonstrate generalizability across linguistic families, we select three languages from different language groups: Spanish (Romance), German (Germanic), and Mandarin (East Asian). The model is fine-tuned on an English (England accent) speech corpus while setting the language identity token to the target base language. During inference, the model receives target-language transcripts and is conditioned on an English reference signal, producing English-accented Spanish, German, or Mandarin speech.

We evaluate English-accent transfer using the objective accent metrics described in Section~\ref{secion:objective_evaluation}, including VoxProfile accent probability and LID-based English (language) probability. In cross-lingual settings where direct accent classifiers are unavailable (e.g., English-accented Spanish), the LID probability serves as a complementary proxy for accent leakage.





\subsubsection{Accent Transfer Results}

Table~\ref{tab:accent_shift_non_english} reports English accent probability (VoxProfile) and English language probability (LID). After fine-tuning, English accent probability increases substantially across all base languages. The LID model similarly assigns higher English language probabilities, providing complementary evidence that the synthesized speech shifts toward English-like acoustic characteristics.

\subsubsection{Intelligibility and Naturalness Analysis}

WER and CER increase in this cross-lingual accent transfer setting. Beyond the general ASR degradation discussed previously, the mismatch is amplified here because language–accent combinations such as English-accented Spanish or English-accented Mandarin are rarely represented in ASR training data. As a result, these samples are more severely out-of-domain than standard accented English, leading to higher recognition errors.

Despite the increase in ASR errors, this configuration achieves relatively strong naturalness scores compared to the non-English-accented English setting (Table~\ref{tab:accent_shift}). This can be attributed to two factors: 1) the fine-tuning corpus (VCTK) has the highest UTMOS among all datasets (Table~\ref{tab:languages_datasets}), and 2) UTMOS is primarily trained on English speech, so shifting acoustic characteristics toward English may align better with its evaluation distribution. Importantly, speaker similarity remains high, indicating that accent transfer does not degrade speaker identity and generalizes effectively across languages.

\subsection{Scaling the Accent Vector}

\begin{figure}[t]
    \centering
     \includegraphics[width=1\linewidth]{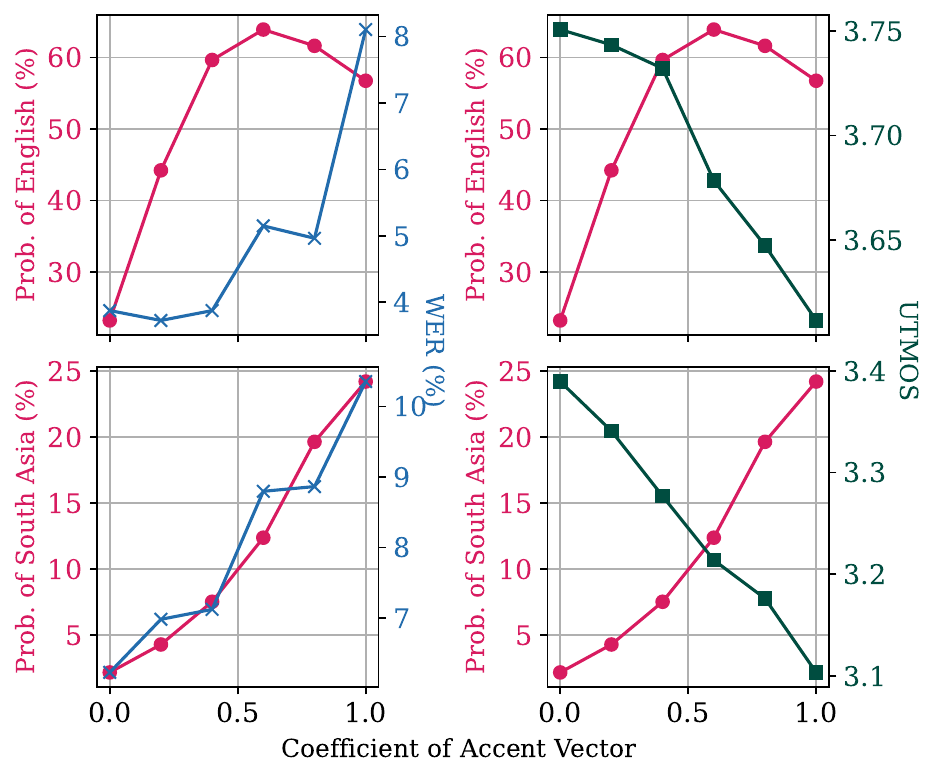}
    \caption{Control of accent strength using the task vector coefficient. We evaluate the model fine-tuned on British and Hindi by measuring accent probability and word error rate (WER) across different task vector coefficients. The first and second rows correspond to the results of British- and Hindi-accented English, respectively. A clear trade-off between accent strength and intelligibility is observed; increasing the coefficient produces a stronger accent but results in higher WER.
    }
    \label{fig:coefficient}
\end{figure}

\subsubsection{Controlling Accent Strength}

To enable controllable accent generation, we adopt the task vector formulation described in Equation~\ref{eq:coefficient}. Accent strength is adjusted by scaling the Accent Vector with a coefficient $\alpha$. We vary $\alpha$ from 0 to 1 in increments of 0.2 to examine controllability and assess the linearity of the Accent Vector space. We evaluate this mechanism in two settings: Hindi-accented English and British-accented English. As shown in Figure~\ref{fig:coefficient}, increasing $\alpha$ results in progressively stronger accent characteristics in both cases. The monotonic trend suggests that accent intensity can be directly and smoothly controlled through linear scaling in the Accent Vector space.

\subsubsection{WER–UTMOS Trade-off with Accent Strength}

Figure~\ref{fig:coefficient} also illustrates the relationship between accent strength and objective quality metrics. As $\alpha$ increases, WER rises while UTMOS gradually declines. This trend reflects a trade-off between accentedness and perceived speech quality under automatic evaluation. Stronger accent realization introduces greater deviation from the ASR model’s training distribution, leading to higher recognition errors, while also slightly reducing predicted naturalness. This behavior is consistent with the ASR mismatch effects discussed in previous sections.


\subsection{Mixed Accent}
\subsubsection{Mixing Multiple Accents}
One notable feature of our method is the ability to mix and merge different accents by manipulating the coefficient of each Accent Vector, following the same principle used for accent strength control in the previous section. In Table~\ref{tab:mixed_accent}, the coefficient of each Accent Vector is set to 0.5, and Equations~\ref{eq:mixed_accent_1} and~\ref{eq:mixed_accent_2} are used to combine two accents. The results show that the probabilities of both target accents generally increase, except in the Spanish + British and Mandarin + British settings. In these cases, the British accent probability increases at the expense of the other accent. We conjecture that this behavior arises because VoxProfile is biased toward English accents, as North American and English accents are a majority in the training data, causing the British accent to dominate when combined with other accents. Furthermore, we observe that the WER is lower for mixed-accent speech than for speech synthesized with a single non-English accent alone. This reduction in WER further emphasizes that the Whisper ASR model exhibits a strong bias toward native English speech patterns, finding the mixed speech more intelligible as its accent shifts closer to that of a native speaker.

\subsubsection{Controlling the Accent Strength of Each Accent}
\begin{figure}
    \centering
    \includegraphics[width=.8\linewidth]{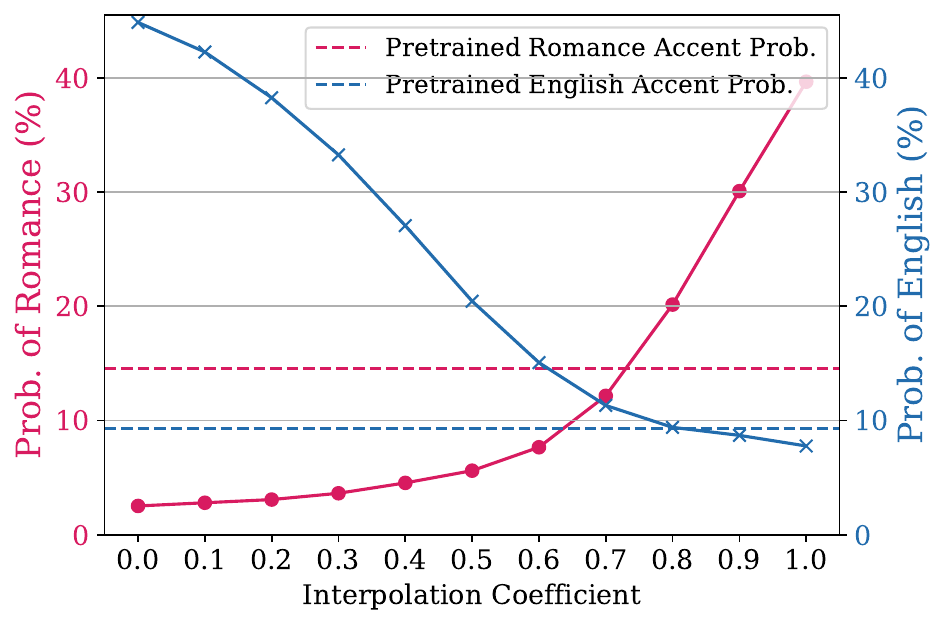}
    \caption{Effect of the task vector coefficient when mixing multiple Accent Vectors. We combine Spanish and English Accent Vectors using coefficients $\alpha$ and $1-\alpha$, respectively. Accent probabilities for Spanish and English accents are evaluated across different values of $\alpha$. The vertical dashed line indicates the accent probability of the pretrained model of each accent.
    }
    \label{fig:mixed_accent}
\end{figure}

Beyond mixing accents, our method also provides fine-grained control over the relative strength of each accent through the corresponding task vector coefficients. We investigate this property using a Spanish–British accent mixing setting. Specifically, the Spanish and British Accent Vectors are combined with coefficients $\alpha$ and $1-\alpha$, respectively. Figure~\ref{fig:mixed_accent} shows that the strength of each accent can be controlled by adjusting the coefficient values. Even in the mixed-accent setting, the coefficients serve as an effective mechanism for controlling individual accent strength.

\begin{table*}[t]
\centering
\caption{Results of interpolating the Accent Vectors. VoxProfile measures accent characteristics, cosine similarity evaluates speaker similarity, and UTMOS. The order of accent probabilities follows the order of the target accent columns. For example, the probabilities for the first row are for Spanish/British. The LoRA weights for both languages are merged with a coefficient of 0.5 each.}
\label{tab:mixed_accent}
\footnotesize
\begin{tabular}{lccccccccc}
\toprule
\multirow{2}{*}{\textbf{Target Accent}}
& \multicolumn{2}{c}{Prob on \textbf{Target Accent} (\%)} 
& \multicolumn{2}{c}{SIM on \textbf{Target Accent}} 
& \multirow{2}{*}{WER / CER}
& \multirow{2}{*}{SSIM} 
& \multirow{2}{*}{MOS} 
 \\
\cmidrule(lr){2-3} \cmidrule(lr){4-5}
& Pretrained
& Finetuned 
& Pretrained 
& Finetuned 
& 
&  
&  \\
\midrule
Spanish + British & 14.58/9.29 & 5.61/20.45 & 0.412/0.295 & 0.307/0.442 & 8.58/5.06 & 0.84 & 3.22  \\
Mandarin + British & 27.38/4.31 & 13.30/17.77 & 0.557/0.156 & 0.445/0.355 & 13.0/9.57 & 0.82 & 3.00   \\
Hindi + British & 2.16/3.39 & 7.11/9.64 & 0.397/0.057 & 0.447/0.157 & 8.24/5.60 & 0.86 & 3.32  \\
\midrule
Spanish + Hindi & 14.58/1.05 & 24.29/10.92 & 0.412/0.271 & 0.561/0.508 & 11.3/7.14 & 0.88 & 3.03   \\
Mandarin + Hindi & 27.38/0.87 & 28.84/8.70 & 0.557/0.316 & 0.617/0.535 & 19.1/13.4 & 0.85 & 2.79  \\
\bottomrule
\end{tabular}
\end{table*}

\subsection{Human Evaluation}

\subsubsection{Accent Identification Accuracy}

\begin{table}[h!]
\vspace{-10pt}
\centering
\caption{Subjective Evaluation Results: Accent Identification Accuracy, Perceived Accent Strength (Correct Responses Only), and Naturalness.}
\label{tab:results_human_eval}
\setlength{\tabcolsep}{4pt}
\footnotesize
\begin{tabular}{l ccc}
\toprule
\textbf{Accent} & \textbf{Accuracy (\%)} & \textbf{Strength} & \textbf{Naturalness} \\
\midrule
US & 80.00 & 3.63 & 3.06 \\
British (England) & 78.46 & 3.55 & 3.91 \\
German & 53.85 & 3.23 & 3.03 \\
French & 66.15 & 3.53 & 3.00 \\
Spanish & 53.85 & 3.00 & 3.28 \\
Hindi & 78.46 & 3.65 & 3.15 \\
Mandarin & 70.77 & 3.13 & 2.31 \\
\bottomrule
\end{tabular}
\end{table}

We compute accent identification accuracy using human-labeled responses against the ground-truth intended accent. American English is used as the baseline, where samples are generated using the pretrained model conditioned on American reference speech. As shown in Table~\ref{tab:results_human_eval}, listeners achieve high accuracy for native English accents, indicating that the baseline speech is clearly recognizable.

Overall, human listeners distinguish accents significantly better than random chance (14\%). While the objective accent classifier reports relatively lower probabilities for some non-English accents (Table~\ref{tab:accent_shift}), human accuracy is substantially higher, suggesting that subjective perception captures accent characteristics beyond what automatic classifiers detect. Furthermore, the perceived accent strength scores indicate that the generated accents are consistently and appropriately noticeable and salient relative to the American English baseline. Naturalness ratings fall between 2 (“slightly natural”) and 4 (“quite natural”), indicating that accent manipulation preserves acceptable perceptual quality.

\subsubsection{Confusion Matrix Analysis}
\begin{figure}[h!]
    \centering
    \vspace{-6mm}
    \includegraphics[width=0.45\textwidth]{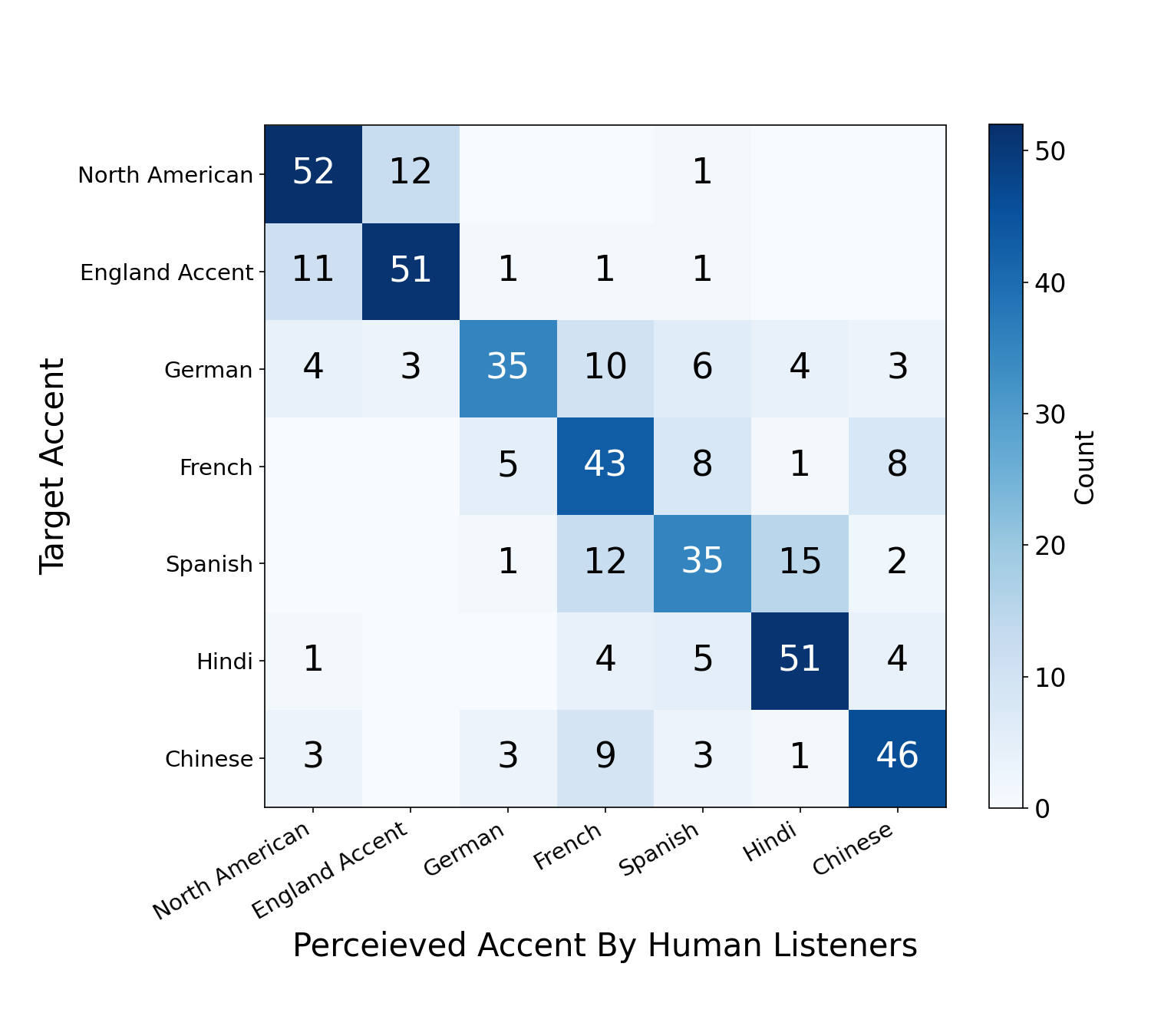}
    \vspace{-5mm}
    \caption{Confusion matrix on the target accents and perceived accents by human listeners.}
    \vspace{-5mm}
    \label{fig:confusion_matrix}
\end{figure}

We construct a confusion matrix using human responses and the corresponding target accents. As shown in Fig.~\ref{fig:confusion_matrix}, listeners most frequently confuse European accents, particularly German-, French-, and Spanish-accented English. This confusion may be influenced by listener bias, as the majority of participants reside in the United States and are not native speakers of these European languages, potentially limiting their sensitivity to subtle inter-accent distinctions.

Overall, the subjective evaluation confirms that the proposed method effectively shifts speech toward the intended target accent, aligning with trends observed in the objective evaluation metrics.

\subsection{Limitation}
While \textit{Accent Vector} demonstrates a practical path to controllable accent manipulation without accented training corpora, several important limitations remain. 

\vspace{1.5mm}
\noindent \textbf{Evaluation Proxies and Domain Mismatch:}
Our objective evaluation relies on pretrained proxies—VoxProfile, an LID model, Whisper ASR, and UTMOS—all of which carry domain biases. These models are predominantly trained on well-represented English varieties and do not adequately cover many cross-lingual accent combinations, leading to potential mismatch effects. Therefore, in our cross-lingual accent transfer setting in particular, absolute automatic scores must be interpreted with extra caution. 

\vspace{1.5mm}
\noindent \textbf{Data and Modeling Constraints:}
The effectiveness of \textit{Accent Vector} depends on both fine-tuning data quality and the linguistic distance between languages. We observed weaker gains for Mandarin, likely due to lower recording quality and substantial prosodic and phonological differences from English. Additionally, representing accent adaptation as an approximately linear parameter shift may be insufficient for capturing complex suprasegmental phenomena, particularly tonal variations. 

\section{Conclusion and Future Work}
We introduced \textit{Accent Vector}, a simple and practical framework for controllable accent manipulation in multilingual TTS that does not require accent-specific speech datasets. By fine-tuning a multilingual backbone with LoRA and extracting task-space parameter differences, \textit{Accent Vector} enable continuous control over accent strength and support linear composition for mixed-accent synthesis. Experiments across multiple languages demonstrate that the approach generalizes beyond English, effectively shifts speech toward target accents, preserves speaker identity, and allows both fine-grained accent control and flexible accent mixing, while revealing interpretable trade-offs between accentedness and automatic intelligibility/naturalness metrics.

\section{Acknowledgement}
\ifcameraready
This work was supported by the Office of the Director of National Intelligence (ODNI), Intelligence Advanced Research Projects Activity (IARPA), via the ARTS Program under contract D2023-2308110001. The views and conclusions contained herein are those of the authors and should not be interpreted as necessarily representing the official policies, either expressed or implied, of ODNI, IARPA, or the U.S. Government. The U.S. Government is authorized to reproduce and distribute reprints for governmental purposes notwithstanding any copyright annotation therein.
\else 
[Hidden for double-blind submission]
\fi

\section{Generative AI Use Disclosure}
Generative AI tools were used solely for language editing, polishing the manuscript, and assisting with writing code. All research ideas, experimental design, implementations, analyses, and reported results were developed and conducted by the authors. The conceptual framing, methodology, and scientific contributions of this work are entirely human-generated.


\bibliographystyle{IEEEtran}
\bibliography{mybib}

\end{document}